\documentclass{article}

\usepackage[preprint]{neurips_2024}

\usepackage[dvipsnames]{xcolor}         
\definecolor{linkColor}{rgb}{0.18,0.39,0.62}
\usepackage[utf8]{inputenc} 
\usepackage[T1]{fontenc}    
\usepackage[colorlinks=true,linkcolor=linkColor,citecolor=linkColor,filecolor=linkColor,urlcolor=linkColor]{hyperref}       

\usepackage{url}            
\usepackage{booktabs}       
\usepackage{amsfonts}       
\usepackage{nicefrac}       
\usepackage{microtype}      
\usepackage{amssymb}

\usepackage{amsmath}
\usepackage{graphicx}
\usepackage{multirow}
\usepackage{pifont}
\usepackage{textcomp}
\usepackage[most]{tcolorbox}
\usepackage{color}
\usepackage{enumitem} 

\newtcolorbox[list inside=prompt]{prompt}[1][]{
    colbacktitle=black!60,
    coltitle=white,
    fontupper=\footnotesize,
    boxsep=5pt,
    left=0pt,
    right=-1pt,
    top=0pt,
    bottom=0pt,
    boxrule=1pt,
    #1,
}

\title{Bootstrap Your Own Context Length}

\author{Liang Wang\thanks{Correspondence to \{wangliang,nanya,fuwei\}@microsoft.com}~~~~Nan Yang~~~Xingxing Zhang~~~Xiaolong Huang~~~Furu Wei\\
Microsoft Corporation \\
{\href{https://aka.ms/GeneralAI}{https://aka.ms/GeneralAI}}
}

\begin{document}

\maketitle

\begin{abstract}

We introduce a bootstrapping approach to train long-context language models
by exploiting their short-context capabilities only.
Our method utilizes a simple agent workflow to synthesize diverse long-context instruction tuning data,
thereby eliminating the necessity for manual data collection and annotation.
The proposed data synthesis workflow requires only a short-context language model,
a text retriever, and a document collection,
all of which are readily accessible within the open-source ecosystem.
Subsequently,
language models are fine-tuned using the synthesized data to extend their context lengths.
In this manner,
we effectively transfer the short-context capabilities of language models to long-context scenarios
through a bootstrapping process.
We conduct experiments with the open-source Llama-3 family of models
and demonstrate that our method can successfully extend the context length to up to 1M tokens,
achieving superior performance across various benchmarks.
\end{abstract}

\section{Introduction}
Long-context large language models (LLM) are essential for understanding long-form texts
across various applications,
including retrieval-augmented generation (RAG) with many documents~\citep{lewis2020retrieval,jiang2024longrag},
repository-level software engineering tasks~\citep{jimenezswe},
and prolonged virtual assistants interactions~\citep{park2023generative}, among others.
Significant advancements has been achieved in training LLMs with increasingly longer context lengths,
ranging from $2$k tokens in LLaMA-1~\citep{touvron2023llama} to $128$k tokens in LLaMA-3~\citep{dubey2024llama},
and even reaching $1$M tokens~\citep{liu2024world,gradientlongcontextllama3}.
Nevertheless,
comprehensive benchmarking~\citep{hsieh2024ruler} reveals that
the performance of current long-context LLMs often drops considerably as the context length increases,
rendering their effective lengths substantially shorter than the claimed lengths.

A critical element in training long-context LLMs is acquiring diverse and high-quality long-context data.
Existing methodologies~\citep{fu2024data,dubey2024llama,pengyarn}
predominantly concentrate on the pre-training phase and rely on filtering long documents from large-scale pre-training corpora
in domains such as books, code repositories, and scientific papers.
As the context length of LLMs surpasses $128$k tokens,
the availability of natural data that can fill the whole context becomes limited,
and the domain diversity of such data is often constrained.

In this paper,
we propose a bootstrapping approach aimed at extending the context length of existing large language models (LLMs)
by leveraging their short-context capabilities.
Our method utilizes a straightforward agent workflow to generate diverse long-context instruction tuning data,
thus obviating the need to rely on the scarce availability of natural long-context data.
It first prompts LLMs to generate diverse instructions,
followed by employing a text retriever to retrieve relevant documents from a large corpus.
For response generation,
a group of query-focused summarization (QFS) agents are recursively applied to
document chunks to filter out irrelevant information
and a response is finally generated from the summaries.
The generated instructions are concatenated with the retrieved documents to form the input,
while the generated response serves as the target output.
In this workflow,
synthesizing a single data point requires multiple LLM inference steps,
yet each step involves processing a short input
that can comfortably fit within the context window of existing LLMs.

Besides extending the maximum input length of LLMs,
we incorporate the idea of instruction back-translation~\citep{li2023self}
to further extend the maximum output length of LLMs.
This technique involves generating synthetic instructions for long documents
and subsequently training LLMs to reconstruct the original documents from the instructions.

We conduct experiments with the open-source Llama-3 family of models
and show that lightweight post-training with our synthetic data can effectively extend the context length to 1M tokens
while maintaining near-perfect performance on the needle-in-haystack task.
On the more challenging RULER benchmark~\citep{hsieh2024ruler},
our model,
\emph{SelfLong-8B-1M},
surpasses other open-source long-context LLMs by a large margin.
Nonetheless,
we still observe a decline in performance as the context length increases,
indicating the necessity for further research to enhance the performance of long-context LLMs.
The trained models are available at \url{https://huggingface.co/self-long}.

\section{Related Work}

\noindent
\textbf{Long-context Language Models}
offers the promise of understanding and generating long-form text,
which is vital for tasks like book-level question answering, repository-level code generation,
multi-document summarization, and more ~\citep{lee2024can,jiang2024longrag}.
Nonetheless,
training these models poses substantial challenges due to
the quadratic computational cost associated with self-attention and the scarcity of long-context data.
One research avenue seeks to extend the context lengths of existing language models
by manipulating the RoPE~\citep{su2024roformer} position embeddings.
For example,
PI~\citep{chen2023extending} employs a linear interpolation of the position ids of the input tokens,
Llama-Long~\citep{xiong2024effective} modifies the base frequency of the RoPE function,
and YaRN~\citep{pengyarn} implements a hybrid approach.
Though these methods achieve decent perplexity scores in a training-free setting,
further fine-tuning is often necessary to enhance long-context performance continually.

Inference with Transformer-based long-context language models can also be both time-consuming and memory-intensive.
MInference~\citep{jiang2024minference} utilizes the sparse attention pattern to speed up the key-value cache prefilling stage,
while RetrievalAttention~\citep{liu2024retrievalattention} reduces generation latency by employing vector search techniques.
However,
inference acceleration often results in performance trade-offs.
This paper concentrates on the development of better long-context language models,
deferring the optimization of inference to future research endeavors.

\noindent
\textbf{Long-context Data Curations}
are pivotal for the training of long-context language models.
Current methodologies~\citep{fu2024data,gao2024train,pengyarn} predominantly rely on the up-sampling of long documents
from large-scale pre-training corpora such as Redpajama~\citep{together2023redpajama} and Fineweb~\citep{penedo2024fineweb}.
Typical sources of long-context data encompass books, scientific papers, and code repositories.
~\citeauthor{fu2024data,gao2024train} find that both the quality and diversity of the data
are crucial for training effective long-context LLMs.
Nevertheless,
naturally occurring long-context data is often scarce and exhibits limited domain diversity.
Another research avenue focuses on generating synthetic long-context data
through methods such as question generation~\citep{an2024make,dubey2024llama},
recursive text summarization~\citep{dubey2024llama},
or document clustering~\citep{gao2024quest}.
Regarding evaluation data,
most benchmarks~\citep{bai2023longbench,shaham2023zeroscrolls} are inadequate for evaluation beyond $128$k tokens.
For evaluation with $1$M context length,
synthetic tasks ~\citep{hsieh2024ruler} are almost exclusively employed.

\noindent
\textbf{Retrieval-Augmented Generation (RAG)}
synergizes the retrieval of relevant documents with LLMs to enhance the factual accuracy of the generated content,
and ensure the incorporation of up-to-date information~\citep{lewis2020retrieval,Karpukhin2020DensePR}.
RAG often necessitates concatenation of multiple retrieved documents to create a long-context input,
even though each individual document is typically short.
This characteristic positions RAG as a crucial application of long-context LLMs.
~\citeauthor{jiang2024longrag,lee2024can} demonstrate that long-context LLMs can ease the demands of retrieval and,
in some instances,
eliminate the need for retrieval entirely.
In this study,
we employ RAG as an approach for synthesizing data to train long-context LLMs.

\section{Method}

\begin{figure*}[ht]
\centering
\includegraphics[width=1.0\textwidth]{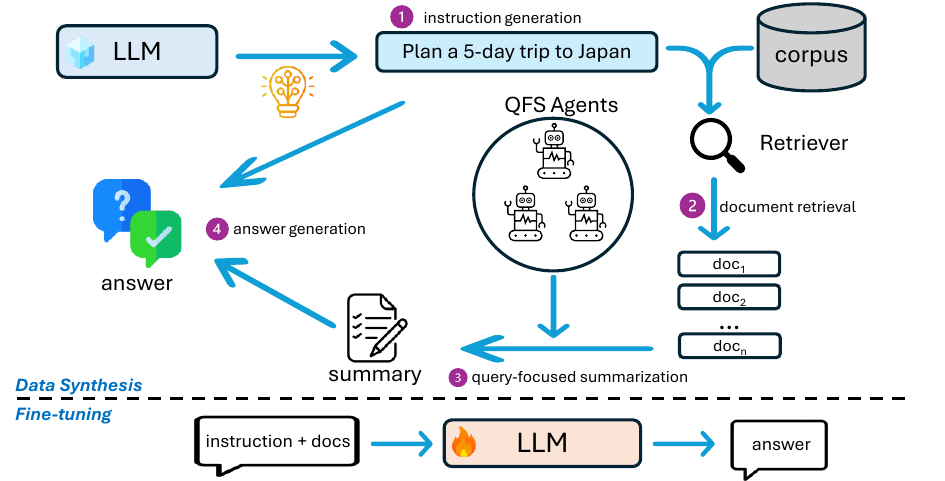}
\caption{The overall workflow for synthesizing long-context instruction tuning data comprises four steps:
instruction generation, relevant document retrieval, recursive query-focused summarization, and response generation.
The generated instructions and retrieved documents are concatenated to form the user-turn input,
whereas the generated response serves as the target output.}
\label{fig:workflow}
\end{figure*}

\subsection{Data Synthesis}

\noindent
\textbf{Long-input Instruction Data via Agent Workflow}
As depicted in Figure~\ref{fig:workflow},
our data synthesis workflow has four steps.
Initially,
an LLM is prompted to generate a diverse array of instructions that require integrating information from multiple documents.
To enhance the diversity of these instructions,
a random text chunk is prepended to each prompt during every LLM call.
This often guides the LLM to generate instructions that are topically relevant to the provided text chunk,
akin to the persona-driven strategy~\citep{ge2024scaling} but simpler.
Next,
an off-the-shelf text retriever,
E5$_\text{mistral-7b}$~\citep{wang2023improving},
is employed to retrieve relevant documents from a large corpus.
The retrieved documents are then split into chunks of at most $4$k tokens
and are fed into a group of query-focused summarization (QFS) agents.
Each QFS agent is tasked with summarizing a document chunk focused on the synthetic instruction,
filtering out information that is irrelevant to the instruction.
This recursive procedure is repeated until the concatenation of all summaries is short enough to be processed by the LLM.
Finally,
the LLM is prompted once more to generate a response based on the summaries and the instruction.

During model training,
the synthetic instruction and retrieved documents are concatenated to form the input,
while the generated response constitutes the target output.
The intermediate summaries are not utilized during training.
The core idea is to decompose the synthesis process into a series of steps,
where each step only requires digesting a short input.
While this particular workflow is selected for its simplicity and effectiveness,
alternative instantiations are also conceivable.

\noindent
\textbf{Long-output Data via Instruction Back-translation}
We first select documents containing between $2$k to $32$k tokens from a high-quality corpus,
and then prompt an LLM to generate a writing instruction that would result in the given document receiving a high evaluation score.
This method of instruction back-translation is inspired by~\cite{li2023self},
although the original work does not focus on long-output generation.

All prompts are provided in the Appendix Section ~\ref{sec:app_prompts}.

\subsection{Training with Long Sequences}
Training with long sequences can be notoriously challenging,
primarily due to the quadratic computational complexity of self-attention,
coupled with the memory constraints of modern accelerators.
To address these challenges,
we employ a progressive training strategy to gradually increase the context length across multiple stages.
At each stage,
we double the maximum context length and
quadruple the RoPE base frequency~\citep{su2024roformer} to ensure a reasonable initialization.
Given that a single H100 with $80$GB of memory can only handle sequences of up to $64$k tokens
for models such as \emph{Llama-3-8B},
even with a batch size of $1$,
we utilize RingAttention~\citep{liu2023ring} to distribute a long input sequence across multiple GPUs.
We perform full-length fine-tuning whenever hardware capabilities allow;
otherwise,
we resort to PoSE-style~\citep{zhu2023pose} training,
which facilitates the decoupling of training length from maximum model length.

When computing the next-token prediction loss,
we average the loss over all input and output tokens for long data samples
to prevent the supervision signal from becoming excessively sparse.
Conversely,
for short-context samples mixed into the training data,
we compute the loss solely over the target output tokens,
disregarding the input tokens.

\section{Experiments}

\subsection{Setup}

\noindent
\textbf{Training Data Mixture }
We combine multiple data sources for training,
including our generated synthetic data and several open-source datasets.
\begin{itemize}[leftmargin=15pt]
\item \textbf{Synthetic Long Instruction Tuning Data} It comprises $69$k long-input samples with $4.6$B tokens, generated based on our proposed agent workflow, and $10$k long-output samples with $77$M tokens, produced via the instruction back-translation method.
\item \textbf{Open-source Instruction Tuning Data} We utilize Tulu-v2~\citep{ivison2023camels} and Infinity-Instruct~\citep{BAAI_Infinity-Instruct} datasets. For Infinity-Instruct, we find that a significant portion of its samples are near-duplicates, so we conduct further de-duplication using the E5$_\text{mistral-7b}$ embedding model.
\item \textbf{Prolong Data}~\citep{gao2024train} is a non-instruction tuning dataset originally employed for the continual pre-training of LLMs. We retain its ``arxiv'', ``book'', ``openwebmath'', ``textbooks'' and ``thestackv1'' portions.
\end{itemize}

The full data mixture encompasses approximately $8.3$B tokens,
with detailed statistics presented in Table ~\ref{tab:app_data_mixture}.
Most samples from the Infinity-Instruct and Tulu-v2 datasets are shorter than $4$k tokens;
thus,
we include them to ensure the model can handle short-context tasks as well.
For synthetic data generation,
we employ \emph{GPT-4o}~\citep{hurst2024gpt} as the backbone LLM
and use the E5$_\text{mistral-7b}$ retriever~\citep{wang2023improving} for document retrieval.
The retrieval corpus contains approximately $10$M documents sampled from the Fineweb-Edu dataset~\citep{penedo2024fineweb}.
We also examine the impact of using \emph{Llama-3.1-8B-Instruct} as backbone LLM in the ablation study.

\noindent
\textbf{Training Procedure }
Our training schedule follows a progressive training strategy.
We start with Llama-3 models that support $128$k tokens and
conduct three sequential stages of training with maximum context lengths of $256$k, $512$k, and $1$M.
At each stage,
we quadruple the RoPE base frequency and switch to PoSE-style efficient training
when a full sequence cannot fit on the hardware.
We apply standard techniques including activation checkpointing, DeepSpeed ZeRO-3, bf16 mixed precision,
FlashAttention~\citep{daoflashattention}, and RingAttention to minimize the memory footprint.
All training is conducted on a single node with 8 H100 GPUs,
while all inference is performed using 8 A100 GPUs.

\noindent
\textbf{Evaluation }
Despite the availability of numerous long-context benchmarks~\citep{bai2023longbench,shaham2023zeroscrolls,zhang2024bench},
the majority are inadequate for evaluation at a context length of $1$M or beyond.
In this study,
we select the RULER benchmark~\citep{hsieh2024ruler},
which comprises $13$ tasks
and allows evaluation at any context length thanks to its automatic data generation process.
To visualize model characteristics at varying depths,
we adopt the needle-in-haystack test,
which requires the model to retrieve a sentence ``\emph{The best thing to do in San Francisco is eat a sandwich and sit in Dolores Park on a sunny day.}''
from a haystack of random essays.
For evaluating long outputs,
we hold out a validation set of $105$ samples ranging from $2$k to $32$k tokens
and compare the model's output length with the groundtruth length.

We utilize vLLM for efficient inference~\citep{kwon2023efficient} across all evaluation tasks.
For \emph{SelfLong-8B-1M},
the prefilling takes about $5$ minutes for a $1$M token sequence,
and the full evaluation on the RULER benchmark takes about $4$ days using $8$ A100 GPUs.

For additional implementation details, please refer to Appendix Section ~\ref{sec:app_implementation}.

\begin{table}[ht]
\centering
\caption{Results on the RULER benchmark spanning context lengths from $32$k to $1$M,
averaged across all $13$ tasks.
The highest and second-highest scores are denoted in bold and underlined, respectively.
Proprietary models are not directly comparable due to the lack of technical details.}
\begin{tabular}{lccccccc}
\toprule
                & Support Length & 32k & 64k & 128k & 256k & 512k & 1M \\ \midrule
FILM-7B  & 128k & 86.9 & 70.1 & 27.1 & - & - & - \\
Phi3-mini  &  128k &  87.5   &  80.6   &  66.7  &  -    &  -    &  -  \\
Llama-3.2-1B-Instruct  & 128k &  64.7  & 43.1  &  0.0  &  -    &  -    &  -  \\
Llama-3.2-3B-Instruct   &  128k &  77.8 &  70.4   &  0.8    &  -    &  -  &   - \\
Llama-3.1-8B-Instruct     & 128k  &  \textbf{89.8}   & \textbf{85.4}  & \underline{78.5}   &  -    &   -   & -   \\
LWM-Text-Chat-1M &  1M &   71.5  &  67.2  &  64.8   &  64.7  &  63.2  &  60.1  \\
Llama-3-8B-1M  &  1M &   81.8  &  78.6  &  77.2   &  \underline{74.2}   &  \underline{70.3}  &  \underline{64.3}  \\ \midrule
SelfLong-1B-1M &  1M  &  61.3 & 56.6  &  54.7  &  46.7  &  40.7  &  31.1 \\
SelfLong-3B-1M &  1M  &  80.5 & 78.0 & 75.5 & 68.8 & 58.5 & 38.8 \\
SelfLong-8B-1M &  1M   &  \underline{89.5} & \underline{84.0} & \textbf{82.0} & \textbf{79.7} & \textbf{78.2} & \textbf{69.6} \\ \midrule \midrule
\multicolumn{8}{l}{\emph{Proprietary models}} \\ \midrule
GPT-4-1106  &  128k &  93.2 &  87.0   &  81.2  &  -    &  -    &  -  \\ \bottomrule
\end{tabular}
\label{tab:main_results}
\end{table}

\subsection{Main Results}

\noindent
\textbf{RULER Benchmark }
We compare against FILM-7B~\citep{an2024make}, Phi3-mini~\citep{abdin2024phi},
the instruct version of Llama-3 models~\citep{dubey2024llama},
LWM-Text-Chat-1M~\citep{liu2024world}, and Llama-3-8B-1M~\citep{gradientlongcontextllama3}.
Additionally,
two proprietary models are included for reference.
Full results of our models can be found in Table ~\ref{tab:app_detailed_ruler}.

The results in Table ~\ref{tab:main_results} reveal several noteworthy observations.
First,
for the official Llama-3 models,
performance markedly declines with reduced model size,
with the 1B and 3B models nearly failing entirely at $128$k context length.
Second,
our models,
initialized from the Llama-3 series,
demonstrate a clear improvement over the official ones,
particularly at $128$k context length.
However,
we do not see consistent performance gain at shorter context lengths,
and a slight decline is occasionally noted.
We hypothesize that a trade-off may exist between varying context lengths given a fixed model capacity,
which warrants further investigation.

\begin{figure*}[ht]
\centering
\includegraphics[width=1.0\textwidth]{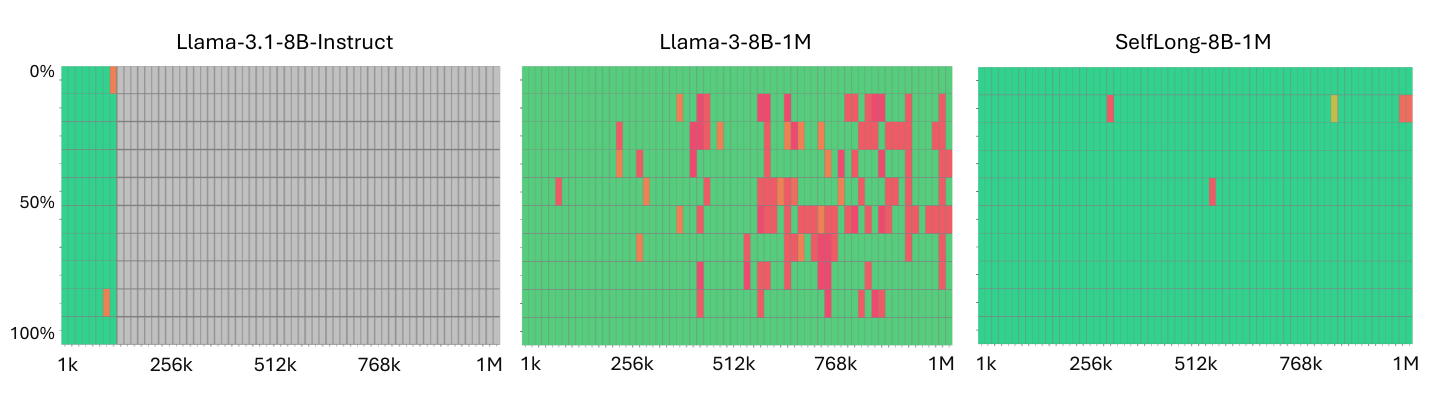}
\caption{Needle-in-haystack test results.
The x-axis represents the context lengths,
while the y-axis indicates the depth of the inserted needle.
The color coding corresponds to the recall score following previous work~\citep{fu2024data},
where green signifies a score close to 1, and red denotes a score close to 0.
A single trial was conducted for each unique combination of context length and needle depth.
The grey shaded regions denote context lengths beyond the model's capability.}
\label{fig:needle_results}
\end{figure*}

\noindent
\textbf{Needle-in-haystack Test }
is a synthetic task designed to assess the capability of LLMs to retrieve a pre-specified needle of varying depth from a long context.
Nonetheless,
the existing literature adopts vastly different evaluation protocols under the same task name.
For instance,
LWM~\citep{liu2024world} utilizes PG19 as the haystack,
with the objective of retrieving a random magic number.
In contrast,
GradientAI~\citep{gradientlongcontextllama3} investigates three different haystacks,
revealing that the performance varies significantly.
In this paper,
we adopt the same evaluation protocol as ~\citeauthor{fu2024data},
which is based on the original one from ~\url{https://github.com/gkamradt/LLMTest_NeedleInAHaystack}.
The needle is a natural language sentence embedded within a haystack of Paul Graham's essays.
For each test,
we calculate the recall score of the needle sentence within the generated text.

Figure ~\ref{fig:needle_results} illustrates that our model achieves near-perfect performance on this task at $1$M context length,
whereas \emph{Llama-3.1-8B-Instruct} is limited to $128$k,
and \emph{Llama-3-8B-1M}~\citep{gradientlongcontextllama3} suffers severe ``lost-in-the-middle'' issues~\citep{liu2024lost}.
A common failure pattern observed is that the model responds based on its own parametric knowledge rather than the provided context.

\begin{figure*}[ht]
\centering
\includegraphics[width=0.9\textwidth]{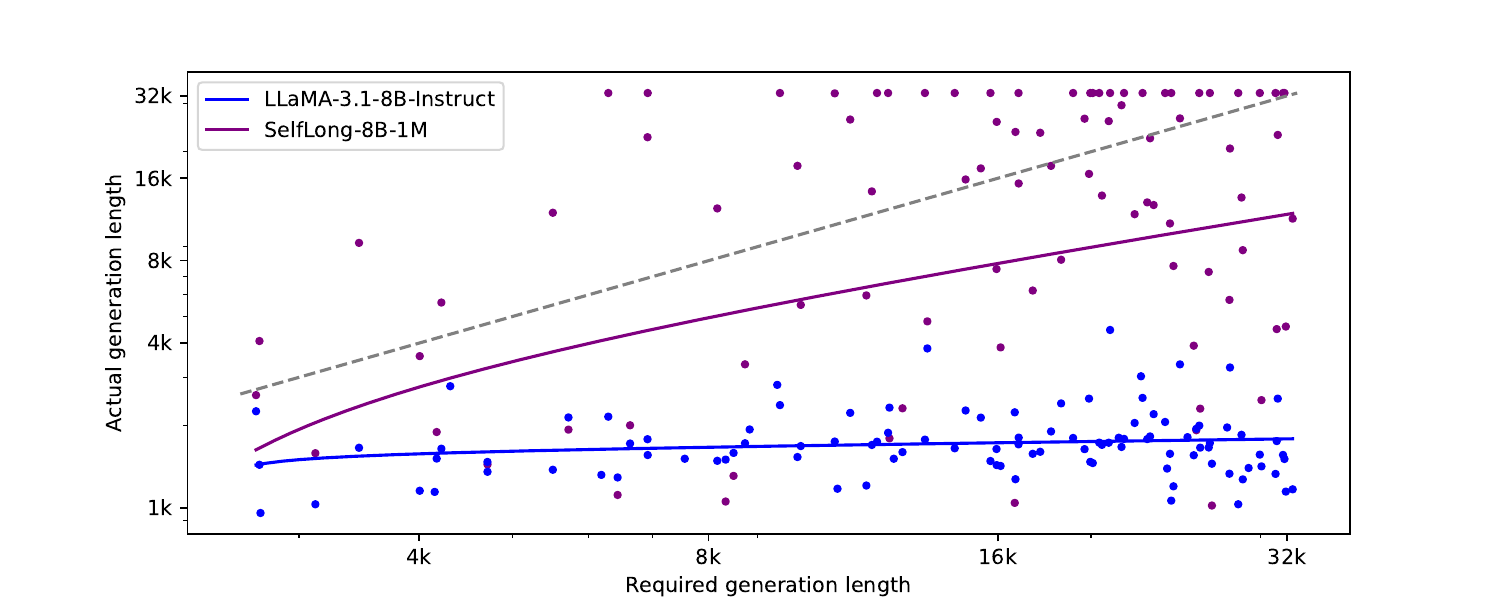}
\caption{Scatter plot illustrating the relationship between the required generation length and the actual output length for samples from the validation set.
The dashed line denotes $y=x$,
indicating the output length precisely matches the groundtruth length.
For each model, we fit a curve to show the trend of the output length as the required length increases.
Details of the curve fitting procedure are provided in Appendix ~\ref{sec:app_implementation}.}
\label{fig:long_output_results}
\end{figure*}

\begin{table}[ht]
\centering
\caption{Average output length for each model on the validation set.
The token count is determined using the Llama-3 tokenizer.}
\begin{tabular}{lccc}
\toprule
                     & Llama-3.1-8B-Instruct & SelfLong-8B-1M & Groundtruth \\ \midrule
Avg. output length       &      1.8k        &       14.5k      &   17.1k          \\ \bottomrule
\end{tabular}
\label{tab:long_output_eval}
\end{table}

\noindent
\textbf{Long Output Generation}
The \emph{Llama-3.1-8B-Instruct} rarely generates outputs exceeding $4$k tokens,
even when the instruction explicitly asks so.
This is substantiated by the data presented in Table ~\ref{tab:long_output_eval} and Figure ~\ref{fig:long_output_results}
based on our held-out validation set.
Our model is able to generate longer outputs,
but its instruction following ability is still imperfect.
As the output length increases,
it frequently deteriorates into repetitive or irrelevant content.
Further research on enhancing and evaluating long output generation represents a promising research avenue.

\section{Analysis}

\subsection{Ablation of Training Recipes}

\begin{figure*}[ht]
\centering
\includegraphics[width=1.0\textwidth]{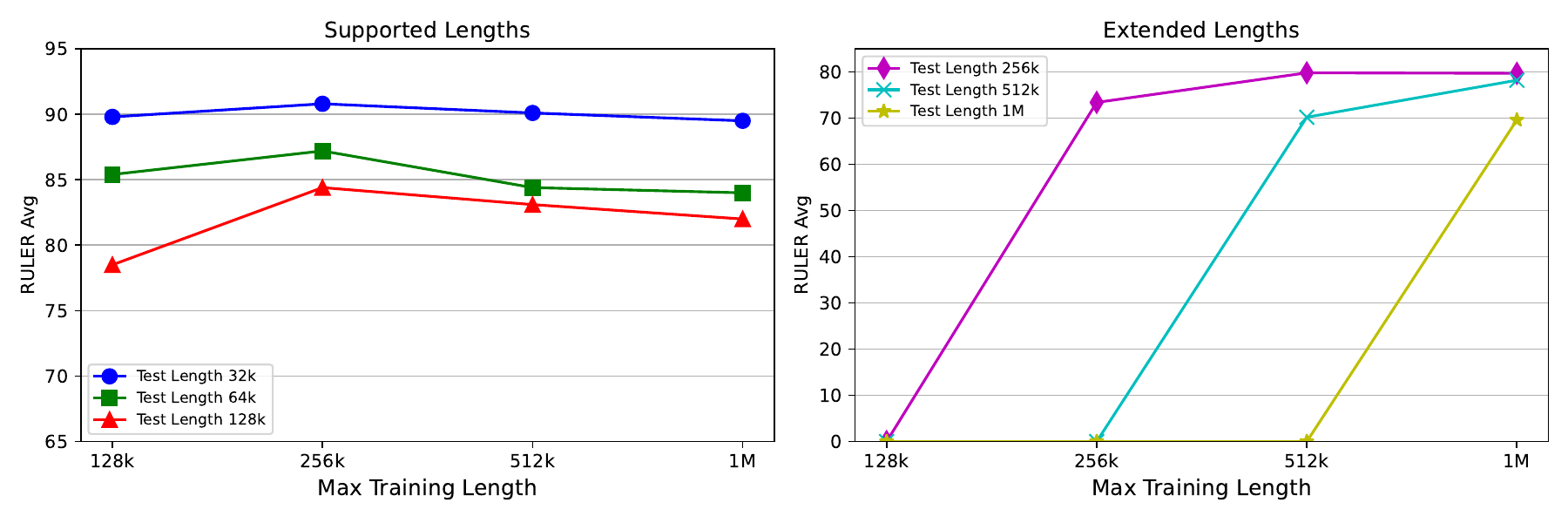}
\caption{The evolving performance across various test lengths as \emph{SelfLong-8B} undergoes progressive training on longer contexts.
The term ``Supported Lengths'' denotes $128$k or shorter,
which \emph{Llama-3.1-8B-Instruct} can already handle.
``Extended Lengths'' refer to the context lengths exceeding $128$k.
If a context length is larger than the model's maximum training length,
the score is assigned a value of $0$.}
\label{fig:progressive}
\end{figure*}

\noindent
\textbf{Effects of Progressive Training }
One research question is how the model's original capability evolves as it is progressively trained on longer contexts
and how the long-context ability emerges.
Figure ~\ref{fig:progressive} shows the evolving scores for each test length.
It is evident that the model's performance on the supported lengths initially improves at $256$k training length,
followed by a gradual decline at $512$k and $1$M.
For lengths exceeding $128$k,
the best performance is achieved when the training length surpasses the test length,
a phenomenon corroborated by previous studies~\citep{gao2024train}.
For instance,
the $256$k score reaches its peak when the training length is $512$k,
rather than $256$k.

In contrast to progressive training,
we also investigate the effects of directly extending to $1$M without intermediate stages.
The results in Table ~\ref{tab:detailed_analysis} indicate a slightly inferior performance relative to the progressive training strategy.
When adopting full-length fine-tuning,
an additional advantage of progressive training is its reduced computational cost
compared to direct extension to maximum length.

\begin{table}[ht]
\centering
\caption{Ablation study results on the RULER benchmark.
The configuration ``w/ adjusted RoPE $\theta$ only'' requires no training data,
whereas ``w/ short data only'' involves fine-tuning on short instruction data with a maximum length of $4$k.
``w/ direct extension to 1M'' directly extends the context length to $1$M without progressive training.
``w/ mask user prompt loss'' masks out all the user prompt tokens during the loss computation.
}
\begin{tabular}{lcccccc}
\toprule
             &   32k   &    64k  & 128k   & 256k   & 512k & 1M \\ \midrule
SelfLong-8B-1M   & \textbf{89.5} & \textbf{84.0} & 82.0 & 79.7 & \textbf{78.2} & \textbf{69.6} \\ \midrule
\multicolumn{7}{l}{\textbf{Data mixture}}           \\
\ \ w/ adjusted RoPE $\theta$ only &   71.0  & 61.2 & 58.8  &   56.1   &  50.6    &  48.1  \\
\ \ w/ short data only  &   83.0   & 69.6  &  61.0 &   55.0   &  48.9   & 46.6   \\
\ \ w/o synthetic data &   84.9 & 79.9 & 78.2  &   77.3   &  72.1   & 63.5  \\ \midrule
\multicolumn{7}{l}{\textbf{Training strategy}}    \\
\ \ w/ direct extension to 1M   &   89.0  & 83.4  & 80.9  &  78.2    &   73.5   &  67.0   \\
\ \ w/ mask user prompt loss   &  88.4   &  83.6  &  \textbf{83.2}  &  \textbf{79.9}    & 76.7   &  66.4 \\ \midrule
\multicolumn{7}{l}{\textbf{LLM for data synthesis}} \\
\ \ w/ Llama-3.1-8B-Instruct   &  83.7  & 81.5  &  80.0    &   77.9     &   73.3   &   66.1 \\ \bottomrule
\end{tabular}
\label{tab:detailed_analysis}
\end{table}

In terms of loss computation,
masking out the user prompt tokens yields a slight performance improvement for context lengths of $128$k and $256$k;
however,
the effects are not consistent across all lengths.
Consequently,
we opt to only mask out the user prompt tokens for short instruction samples
to incorporate more supervision signals.

\noindent
\textbf{Choice of LLMs for Data Synthesis }
In this paper,
we employ \emph{GPT-4o} as the backbone LLM for data synthesis.
To fully explore the idea of bootstrapping on its own,
we also investigate the impact of using \emph{Llama-3.1-8B-Instruct} itself as the backbone.
As illustrated in Table ~\ref{tab:detailed_analysis},
the configuration ``w/ Llama-3.1-8B-Instruct'' demonstrates a decent performance at longer context lengths;
however,
a performance gap remains when compared to using \emph{GPT-4o}.

\noindent
\textbf{Is Training on Short-Context Data Sufficient?}
Due to the scarcity of long-context instruction tuning datasets,
numerous existing studies~\citep{gao2024train,gradientlongcontextllama3} exclusively fine-tune on short instruction data.
When fine-tuned with a maximum length of $4$k,
the model is able to surpass its initialization ``w/ adjusted RoPE $\theta$ only'' within $128$k context length
as presented in Table ~\ref{tab:detailed_analysis}.
Nevertheless,
the model's generalization to longer contexts falls short,
exhibiting even poorer performance compared to ``w/ adjusted RoPE $\theta$ only'',
where no training is performed.
This preliminary finding underscores the necessity for
curating high-quality long-context instruction data for LLM post-training.

\subsection{Extending to 4M Context Length}

\begin{figure*}[ht]
\centering
\includegraphics[width=1.0\textwidth]{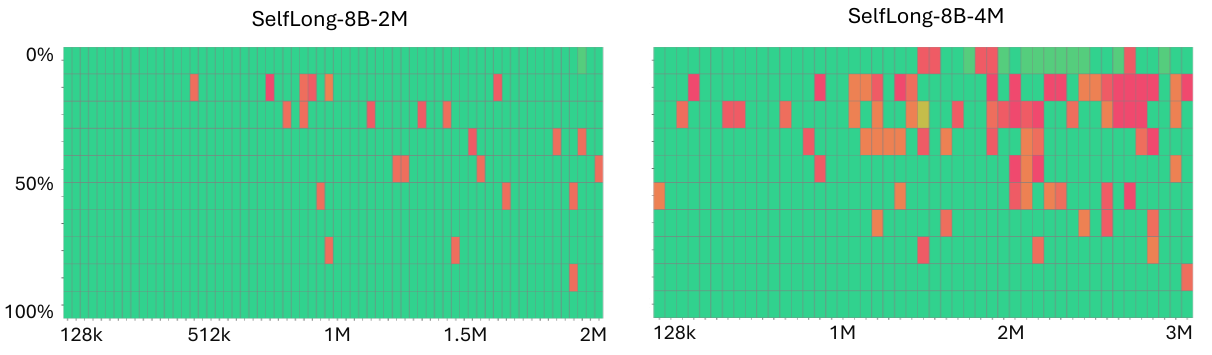}
\caption{Needle-in-haystack test results when extending the context length up to 4M.
For the $4$M version,
tests were conducted within $3$M context length due to the prohibitively high inference cost.}
\label{fig:4m_needle}
\end{figure*}

To test the limits of long-context modeling within academic compute budgets (8 H100 GPUs),
we further extend the context length to $4$M tokens
through two additional progressive training stages.
The needle-in-haystack test results,
as depicted in Figure ~\ref{fig:4m_needle},
indicate that this simple test becomes increasingly challenging as the context length increases.
A further complication arises from the exceedingly high inference cost;
for instance,
prefilling a single $3$M token sequence requires approximately $30$ minutes for an 8B model,
while the key-value cache demands about $400$GB of GPU memory.
This necessitates advancements in model architecture~\citep{sun2024you,ding2023longnet} and system optimization
to make long-context LLMs more affordable.

\subsection{Performance on Short-Context Tasks}

\begin{table}[ht]
\centering
\caption{Performance on short-context tasks from the Open LLM Leaderboard 2 before and after our fine-tuning.
We use the official metrics from \emph{lm-evaluation-harness}~\citep{eval-harness} for all tasks.}
\begin{tabular}{lccccc}
\toprule
 & BBH & GPQA & IFEval & MMLU Pro & MUSR \\ \midrule
Llama-3.2-3B-Instruct & 45.9  & 27.9  &  73.9 & 31.9  & 35.2  \\
SelfLong-3B-1M  & 42.4 & 27.6 & 57.0  & 28.6  & 39.5  \\ \midrule \midrule
Llama-3.1-8B-Instruct & 50.1  & 26.8 & 77.4 & 37.5 & 36.9 \\
SelfLong-8B-1M & 48.7  & 30.4 & 65.9  & 35.3  & 41.5  \\ \bottomrule
\end{tabular}
\label{tab:short_context_tasks}
\end{table}

In addition to long-context tasks,
we also evaluate our models on tasks from the Open LLM Leaderboard 2,
which includes BBH~\citep{suzgun2022challenging}, GPQA~\citep{rein2023gpqa},
IFEval~\citep{zhou2023instructionfollowing}, MMLU Pro~\citep{wang2024mmluprorobustchallengingmultitask},
and MUSR~\citep{sprague2024musrtestinglimitschainofthought}.
After context extension,
our models maintain competitive scores on these short-context tasks,
as illustrated in Table ~\ref{tab:short_context_tasks},
with one exception of the IFEval task.
We hypothesize that using a better post-training data mixture,
such as Tulu-3~\citep{lambert2024t},
could help mitigate the performance loss on IFEval.

\subsection{Solving Long-Context Tasks with Agent Workflow}

\begin{table}[ht]
\centering
\caption{Comparison of solving long-context tasks using agent workflow versus long-context LLM at $128$k length.
Both approaches utilize \emph{Llama-3.1-8B-Instruct} as the backbone LLM.}
\begin{tabular}{lccccc}
\toprule
                 &  Avg. \# of LLM calls & niah\_single\_1 & vt & qa\_1 & qa\_2 \\ \midrule
Long-context LLM &  1 &   100   &  \textbf{58.2}  &   80.0    &    47.0   \\
Agent Workflow   &  33  &  100   &  25.4  &   \textbf{89.0}    &   \textbf{59.0}    \\ \bottomrule
\end{tabular}
\label{tab:ablation_long_context_or_agent}
\end{table}

Our agent workflow for data synthesis offers an alternative approach for solving long-context tasks.
Rather than feeding the entire context into the model,
we can break down the long context into multiple chunks,
employing the workflow in Figure ~\ref{fig:workflow} to generate the answer.
A $128$k-length context is split into $32$ chunks of $4$k tokens each,
which are then treated as the retrieved documents within the agent workflow.

Table ~\ref{tab:ablation_long_context_or_agent} presents the results for several representative tasks
from the RULER benchmark.
Our analysis reveals that long-context LLM and agent workflow exhibit complementary strengths.
The agent workflow excels at QA tasks that require collecting small pieces of relevant information from a long context.
However,
it encounters difficulties with tasks requiring sequential reasoning throughout the entire context,
such as the Variable Tracking (vt) task.
To solve one task instance,
agent workflow requires significantly more LLM calls,
though each call is much cheaper due to the shorter context processed.
Future research could explore the potential of integrating these two methods
from the perspective of inference efficiency and task performance.

\section{Conclusion}
This paper presents an effective recipe to extend the context length of existing LLMs
by leveraging their short-context capabilities to synthesize long instruction tuning data.
Our proposed data synthesis framework involves the collaboration of
multiple agents and a document retriever to generate diverse long-context data
through multiple inference steps.
Experiments with the open-source Llama-3 models demonstrate that
our approach successfully extends the context length to $1$M tokens,
achieving competitive performance across a range of long-context tasks.
Future work includes developing more effective data synthesis workflows,
improving the inference efficiency of long-context LLMs,
and exploring the potential of long-context LLMs in real-world applications.

\bibliography{custom}

\begin{thebibliography}{47}
\providecommand{\natexlab}[1]{#1}
\providecommand{\url}[1]{\texttt{#1}}
\expandafter\ifx\csname urlstyle\endcsname\relax
  \providecommand{\doi}[1]{doi: #1}\else
  \providecommand{\doi}{doi: \begingroup \urlstyle{rm}\Url}\fi

\bibitem[Abdin et~al.(2024)Abdin, Aneja, Awadalla, Awadallah, Awan, Bach, Bahree, Bakhtiari, Bao, Behl, et~al.]{abdin2024phi}
Marah Abdin, Jyoti Aneja, Hany Awadalla, Ahmed Awadallah, Ammar~Ahmad Awan, Nguyen Bach, Amit Bahree, Arash Bakhtiari, Jianmin Bao, Harkirat Behl, et~al.
\newblock Phi-3 technical report: A highly capable language model locally on your phone.
\newblock \emph{ArXiv preprint}, abs/2404.14219, 2024.
\newblock URL \url{https://arxiv.org/abs/2404.14219}.

\bibitem[An et~al.(2024)An, Ma, Lin, Zheng, and Lou]{an2024make}
Shengnan An, Zexiong Ma, Zeqi Lin, Nanning Zheng, and Jian-Guang Lou.
\newblock Make your llm fully utilize the context.
\newblock \emph{ArXiv preprint}, abs/2404.16811, 2024.
\newblock URL \url{https://arxiv.org/abs/2404.16811}.

\bibitem[BAAI()]{BAAI_Infinity-Instruct}
BAAI.
\newblock Infinity-instruct.
\newblock URL \url{https://huggingface.co/datasets/BAAI/Infinity-Instruct}.

\bibitem[Bai et~al.(2023)Bai, Lv, Zhang, Lyu, Tang, Huang, Du, Liu, Zeng, Hou, et~al.]{bai2023longbench}
Yushi Bai, Xin Lv, Jiajie Zhang, Hongchang Lyu, Jiankai Tang, Zhidian Huang, Zhengxiao Du, Xiao Liu, Aohan Zeng, Lei Hou, et~al.
\newblock Longbench: A bilingual, multitask benchmark for long context understanding.
\newblock \emph{ArXiv preprint}, abs/2308.14508, 2023.
\newblock URL \url{https://arxiv.org/abs/2308.14508}.

\bibitem[Chen et~al.(2023)Chen, Wong, Chen, and Tian]{chen2023extending}
Shouyuan Chen, Sherman Wong, Liangjian Chen, and Yuandong Tian.
\newblock Extending context window of large language models via positional interpolation.
\newblock \emph{ArXiv preprint}, abs/2306.15595, 2023.
\newblock URL \url{https://arxiv.org/abs/2306.15595}.

\bibitem[Computer(2023)]{together2023redpajama}
Together Computer.
\newblock Redpajama: an open dataset for training large language models, 2023.
\newblock URL \url{https://github.com/togethercomputer/RedPajama-Data}.

\bibitem[Dao(2024)]{daoflashattention}
Tri Dao.
\newblock Flashattention-2: Faster attention with better parallelism and work partitioning.
\newblock In \emph{The Twelfth International Conference on Learning Representations, {ICLR} 2024, Vienna, Austria, May 7-11, 2024}. OpenReview.net, 2024.
\newblock URL \url{https://openreview.net/forum?id=mZn2Xyh9Ec}.

\bibitem[Ding et~al.(2023)Ding, Ma, Dong, Zhang, Huang, Wang, Zheng, and Wei]{ding2023longnet}
Jiayu Ding, Shuming Ma, Li~Dong, Xingxing Zhang, Shaohan Huang, Wenhui Wang, Nanning Zheng, and Furu Wei.
\newblock Longnet: Scaling transformers to 1,000,000,000 tokens.
\newblock \emph{ArXiv preprint}, abs/2307.02486, 2023.
\newblock URL \url{https://arxiv.org/abs/2307.02486}.

\bibitem[Dubey et~al.(2024)Dubey, Jauhri, Pandey, Kadian, Al-Dahle, Letman, Mathur, Schelten, Yang, Fan, et~al.]{dubey2024llama}
Abhimanyu Dubey, Abhinav Jauhri, Abhinav Pandey, Abhishek Kadian, Ahmad Al-Dahle, Aiesha Letman, Akhil Mathur, Alan Schelten, Amy Yang, Angela Fan, et~al.
\newblock The llama 3 herd of models.
\newblock \emph{ArXiv preprint}, abs/2407.21783, 2024.
\newblock URL \url{https://arxiv.org/abs/2407.21783}.

\bibitem[Fu et~al.(2024)Fu, Panda, Niu, Yue, Hajishirzi, Kim, and Peng]{fu2024data}
Yao Fu, Rameswar Panda, Xinyao Niu, Xiang Yue, Hannaneh Hajishirzi, Yoon Kim, and Hao Peng.
\newblock Data engineering for scaling language models to 128k context.
\newblock In \emph{Forty-first International Conference on Machine Learning, {ICML} 2024, Vienna, Austria, July 21-27, 2024}. OpenReview.net, 2024.
\newblock URL \url{https://openreview.net/forum?id=TaAqeo7lUh}.

\bibitem[Gao et~al.(2024{\natexlab{a}})Gao, Wu, Fu, and Hu]{gao2024quest}
Chaochen Gao, Xing Wu, Qi~Fu, and Songlin Hu.
\newblock Quest: Query-centric data synthesis approach for long-context scaling of large language model.
\newblock \emph{ArXiv preprint}, abs/2405.19846, 2024{\natexlab{a}}.
\newblock URL \url{https://arxiv.org/abs/2405.19846}.

\bibitem[Gao et~al.(2024{\natexlab{b}})Gao, Tow, Abbasi, Biderman, Black, DiPofi, Foster, Golding, Hsu, Le~Noac'h, Li, McDonell, Muennighoff, Ociepa, Phang, Reynolds, Schoelkopf, Skowron, Sutawika, Tang, Thite, Wang, Wang, and Zou]{eval-harness}
Leo Gao, Jonathan Tow, Baber Abbasi, Stella Biderman, Sid Black, Anthony DiPofi, Charles Foster, Laurence Golding, Jeffrey Hsu, Alain Le~Noac'h, Haonan Li, Kyle McDonell, Niklas Muennighoff, Chris Ociepa, Jason Phang, Laria Reynolds, Hailey Schoelkopf, Aviya Skowron, Lintang Sutawika, Eric Tang, Anish Thite, Ben Wang, Kevin Wang, and Andy Zou.
\newblock A framework for few-shot language model evaluation, 2024{\natexlab{b}}.
\newblock URL \url{https://zenodo.org/records/12608602}.

\bibitem[Gao et~al.(2024{\natexlab{c}})Gao, Wettig, Yen, and Chen]{gao2024train}
Tianyu Gao, Alexander Wettig, Howard Yen, and Danqi Chen.
\newblock How to train long-context language models (effectively).
\newblock \emph{ArXiv preprint}, abs/2410.02660, 2024{\natexlab{c}}.
\newblock URL \url{https://arxiv.org/abs/2410.02660}.

\bibitem[Ge et~al.(2024)Ge, Chan, Wang, Yu, Mi, and Yu]{ge2024scaling}
Tao Ge, Xin Chan, Xiaoyang Wang, Dian Yu, Haitao Mi, and Dong Yu.
\newblock Scaling synthetic data creation with 1,000,000,000 personas.
\newblock \emph{ArXiv preprint}, abs/2406.20094, 2024.
\newblock URL \url{https://arxiv.org/abs/2406.20094}.

\bibitem[Hsieh et~al.(2024)Hsieh, Sun, Kriman, Acharya, Rekesh, Jia, Zhang, and Ginsburg]{hsieh2024ruler}
Cheng-Ping Hsieh, Simeng Sun, Samuel Kriman, Shantanu Acharya, Dima Rekesh, Fei Jia, Yang Zhang, and Boris Ginsburg.
\newblock Ruler: What's the real context size of your long-context language models?
\newblock \emph{ArXiv preprint}, abs/2404.06654, 2024.
\newblock URL \url{https://arxiv.org/abs/2404.06654}.

\bibitem[Hurst et~al.(2024)Hurst, Lerer, Goucher, Perelman, Ramesh, Clark, Ostrow, Welihinda, Hayes, Radford, et~al.]{hurst2024gpt}
Aaron Hurst, Adam Lerer, Adam~P Goucher, Adam Perelman, Aditya Ramesh, Aidan Clark, AJ~Ostrow, Akila Welihinda, Alan Hayes, Alec Radford, et~al.
\newblock Gpt-4o system card.
\newblock \emph{ArXiv preprint}, abs/2410.21276, 2024.
\newblock URL \url{https://arxiv.org/abs/2410.21276}.

\bibitem[Ivison et~al.(2023)Ivison, Wang, Pyatkin, Lambert, Peters, Dasigi, Jang, Wadden, Smith, Beltagy, et~al.]{ivison2023camels}
Hamish Ivison, Yizhong Wang, Valentina Pyatkin, Nathan Lambert, Matthew Peters, Pradeep Dasigi, Joel Jang, David Wadden, Noah~A Smith, Iz~Beltagy, et~al.
\newblock Camels in a changing climate: Enhancing lm adaptation with tulu 2.
\newblock \emph{ArXiv preprint}, abs/2311.10702, 2023.
\newblock URL \url{https://arxiv.org/abs/2311.10702}.

\bibitem[Jiang et~al.(2024{\natexlab{a}})Jiang, Li, Zhang, Wu, Luo, Ahn, Han, Abdi, Li, Lin, et~al.]{jiang2024minference}
Huiqiang Jiang, Yucheng Li, Chengruidong Zhang, Qianhui Wu, Xufang Luo, Surin Ahn, Zhenhua Han, Amir~H Abdi, Dongsheng Li, Chin-Yew Lin, et~al.
\newblock Minference 1.0: Accelerating pre-filling for long-context llms via dynamic sparse attention.
\newblock \emph{ArXiv preprint}, abs/2407.02490, 2024{\natexlab{a}}.
\newblock URL \url{https://arxiv.org/abs/2407.02490}.

\bibitem[Jiang et~al.(2024{\natexlab{b}})Jiang, Ma, and Chen]{jiang2024longrag}
Ziyan Jiang, Xueguang Ma, and Wenhu Chen.
\newblock Longrag: Enhancing retrieval-augmented generation with long-context llms.
\newblock \emph{ArXiv preprint}, abs/2406.15319, 2024{\natexlab{b}}.
\newblock URL \url{https://arxiv.org/abs/2406.15319}.

\bibitem[Jimenez et~al.(2024)Jimenez, Yang, Wettig, Yao, Pei, Press, and Narasimhan]{jimenezswe}
Carlos~E. Jimenez, John Yang, Alexander Wettig, Shunyu Yao, Kexin Pei, Ofir Press, and Karthik~R. Narasimhan.
\newblock Swe-bench: Can language models resolve real-world github issues?
\newblock In \emph{The Twelfth International Conference on Learning Representations, {ICLR} 2024, Vienna, Austria, May 7-11, 2024}. OpenReview.net, 2024.
\newblock URL \url{https://openreview.net/forum?id=VTF8yNQM66}.

\bibitem[Karpukhin et~al.(2020)Karpukhin, Oguz, Min, Lewis, Wu, Edunov, Chen, and Yih]{Karpukhin2020DensePR}
Vladimir Karpukhin, Barlas Oguz, Sewon Min, Patrick Lewis, Ledell Wu, Sergey Edunov, Danqi Chen, and Wen-tau Yih.
\newblock Dense passage retrieval for open-domain question answering.
\newblock In Bonnie Webber, Trevor Cohn, Yulan He, and Yang Liu, editors, \emph{Proceedings of the 2020 Conference on Empirical Methods in Natural Language Processing (EMNLP)}, pages 6769--6781, Online, 2020. Association for Computational Linguistics.
\newblock \doi{10.18653/v1/2020.emnlp-main.550}.
\newblock URL \url{https://aclanthology.org/2020.emnlp-main.550}.

\bibitem[Kwon et~al.(2023)Kwon, Li, Zhuang, Sheng, Zheng, Yu, Gonzalez, Zhang, and Stoica]{kwon2023efficient}
Woosuk Kwon, Zhuohan Li, Siyuan Zhuang, Ying Sheng, Lianmin Zheng, Cody~Hao Yu, Joseph Gonzalez, Hao Zhang, and Ion Stoica.
\newblock Efficient memory management for large language model serving with pagedattention.
\newblock In \emph{Proceedings of the 29th Symposium on Operating Systems Principles}, pages 611--626, 2023.

\bibitem[Lambert et~al.(2024)Lambert, Morrison, Pyatkin, Huang, Ivison, Brahman, Miranda, Liu, Dziri, Lyu, et~al.]{lambert2024t}
Nathan Lambert, Jacob Morrison, Valentina Pyatkin, Shengyi Huang, Hamish Ivison, Faeze Brahman, Lester James~V Miranda, Alisa Liu, Nouha Dziri, Shane Lyu, et~al.
\newblock T$\backslash$" ulu 3: Pushing frontiers in open language model post-training.
\newblock \emph{ArXiv preprint}, abs/2411.15124, 2024.
\newblock URL \url{https://arxiv.org/abs/2411.15124}.

\bibitem[Lee et~al.(2024)Lee, Chen, Dai, Dua, Sachan, Boratko, Luan, Arnold, Perot, Dalmia, et~al.]{lee2024can}
Jinhyuk Lee, Anthony Chen, Zhuyun Dai, Dheeru Dua, Devendra~Singh Sachan, Michael Boratko, Yi~Luan, S{\'e}bastien~MR Arnold, Vincent Perot, Siddharth Dalmia, et~al.
\newblock Can long-context language models subsume retrieval, rag, sql, and more?
\newblock \emph{ArXiv preprint}, abs/2406.13121, 2024.
\newblock URL \url{https://arxiv.org/abs/2406.13121}.

\bibitem[Lewis et~al.(2020)Lewis, Perez, Piktus, Petroni, Karpukhin, Goyal, K{\"{u}}ttler, Lewis, Yih, Rockt{\"{a}}schel, Riedel, and Kiela]{lewis2020retrieval}
Patrick S.~H. Lewis, Ethan Perez, Aleksandra Piktus, Fabio Petroni, Vladimir Karpukhin, Naman Goyal, Heinrich K{\"{u}}ttler, Mike Lewis, Wen{-}tau Yih, Tim Rockt{\"{a}}schel, Sebastian Riedel, and Douwe Kiela.
\newblock Retrieval-augmented generation for knowledge-intensive {NLP} tasks.
\newblock In Hugo Larochelle, Marc'Aurelio Ranzato, Raia Hadsell, Maria{-}Florina Balcan, and Hsuan{-}Tien Lin, editors, \emph{Advances in Neural Information Processing Systems 33: Annual Conference on Neural Information Processing Systems 2020, NeurIPS 2020, December 6-12, 2020, virtual}, 2020.
\newblock URL \url{https://proceedings.neurips.cc/paper/2020/hash/6b493230205f780e1bc26945df7481e5-Abstract.html}.

\bibitem[Li et~al.(2024)Li, Yu, Zhou, Schick, Levy, Zettlemoyer, Weston, and Lewis]{li2023self}
Xian Li, Ping Yu, Chunting Zhou, Timo Schick, Omer Levy, Luke Zettlemoyer, Jason Weston, and Mike Lewis.
\newblock Self-alignment with instruction backtranslation.
\newblock In \emph{The Twelfth International Conference on Learning Representations, {ICLR} 2024, Vienna, Austria, May 7-11, 2024}. OpenReview.net, 2024.
\newblock URL \url{https://openreview.net/forum?id=1oijHJBRsT}.

\bibitem[Liu et~al.(2024{\natexlab{a}})Liu, Chen, Lu, Jiang, Han, Zhang, Chen, Zhang, Ding, Zhang, et~al.]{liu2024retrievalattention}
Di~Liu, Meng Chen, Baotong Lu, Huiqiang Jiang, Zhenhua Han, Qianxi Zhang, Qi~Chen, Chengruidong Zhang, Bailu Ding, Kai Zhang, et~al.
\newblock Retrievalattention: Accelerating long-context llm inference via vector retrieval.
\newblock \emph{ArXiv preprint}, abs/2409.10516, 2024{\natexlab{a}}.
\newblock URL \url{https://arxiv.org/abs/2409.10516}.

\bibitem[Liu et~al.(2024{\natexlab{b}})Liu, Yan, Zaharia, and Abbeel]{liu2024world}
Hao Liu, Wilson Yan, Matei Zaharia, and Pieter Abbeel.
\newblock World model on million-length video and language with ringattention.
\newblock \emph{ArXiv preprint}, abs/2402.08268, 2024{\natexlab{b}}.
\newblock URL \url{https://arxiv.org/abs/2402.08268}.

\bibitem[Liu et~al.(2024{\natexlab{c}})Liu, Zaharia, and Abbeel]{liu2023ring}
Hao Liu, Matei Zaharia, and Pieter Abbeel.
\newblock Ringattention with blockwise transformers for near-infinite context.
\newblock In \emph{The Twelfth International Conference on Learning Representations, {ICLR} 2024, Vienna, Austria, May 7-11, 2024}. OpenReview.net, 2024{\natexlab{c}}.
\newblock URL \url{https://openreview.net/forum?id=WsRHpHH4s0}.

\bibitem[Liu et~al.(2024{\natexlab{d}})Liu, Lin, Hewitt, Paranjape, Bevilacqua, Petroni, and Liang]{liu2024lost}
Nelson~F. Liu, Kevin Lin, John Hewitt, Ashwin Paranjape, Michele Bevilacqua, Fabio Petroni, and Percy Liang.
\newblock Lost in the middle: How language models use long contexts.
\newblock \emph{Transactions of the Association for Computational Linguistics}, 12:\penalty0 157--173, 2024{\natexlab{d}}.
\newblock \doi{10.1162/tacl_a_00638}.
\newblock URL \url{https://aclanthology.org/2024.tacl-1.9}.

\bibitem[Park et~al.(2023)Park, O'Brien, Cai, Morris, Liang, and Bernstein]{park2023generative}
Joon~Sung Park, Joseph O'Brien, Carrie~Jun Cai, Meredith~Ringel Morris, Percy Liang, and Michael~S Bernstein.
\newblock Generative agents: Interactive simulacra of human behavior.
\newblock In \emph{Proceedings of the 36th annual acm symposium on user interface software and technology}, pages 1--22, 2023.

\bibitem[Pekelis et~al.(2024)Pekelis, Feil, Moret, Huang, and Peng]{gradientlongcontextllama3}
Leonid Pekelis, Michael Feil, Forrest Moret, Mark Huang, and Tiffany Peng.
\newblock Llama 3 gradient: A series of long context models, 2024.
\newblock URL \url{https://gradient.ai/blog/scaling-rotational-embeddings-for-long-context-language-models}.

\bibitem[Penedo et~al.(2024)Penedo, Kydl{\'\i}{\v{c}}ek, Lozhkov, Mitchell, Raffel, Von~Werra, Wolf, et~al.]{penedo2024fineweb}
Guilherme Penedo, Hynek Kydl{\'\i}{\v{c}}ek, Anton Lozhkov, Margaret Mitchell, Colin Raffel, Leandro Von~Werra, Thomas Wolf, et~al.
\newblock The fineweb datasets: Decanting the web for the finest text data at scale.
\newblock \emph{ArXiv preprint}, abs/2406.17557, 2024.
\newblock URL \url{https://arxiv.org/abs/2406.17557}.

\bibitem[Peng et~al.(2024)Peng, Quesnelle, Fan, and Shippole]{pengyarn}
Bowen Peng, Jeffrey Quesnelle, Honglu Fan, and Enrico Shippole.
\newblock Yarn: Efficient context window extension of large language models.
\newblock In \emph{The Twelfth International Conference on Learning Representations, {ICLR} 2024, Vienna, Austria, May 7-11, 2024}. OpenReview.net, 2024.
\newblock URL \url{https://openreview.net/forum?id=wHBfxhZu1u}.

\bibitem[Rein et~al.(2023)Rein, Hou, Stickland, Petty, Pang, Dirani, Michael, and Bowman]{rein2023gpqa}
David Rein, Betty~Li Hou, Asa~Cooper Stickland, Jackson Petty, Richard~Yuanzhe Pang, Julien Dirani, Julian Michael, and Samuel~R. Bowman.
\newblock Gpqa: A graduate-level google-proof q\&a benchmark, 2023.

\bibitem[Shaham et~al.(2023)Shaham, Ivgi, Efrat, Berant, and Levy]{shaham2023zeroscrolls}
Uri Shaham, Maor Ivgi, Avia Efrat, Jonathan Berant, and Omer Levy.
\newblock {Z}ero{SCROLLS}: A zero-shot benchmark for long text understanding.
\newblock In Houda Bouamor, Juan Pino, and Kalika Bali, editors, \emph{Findings of the Association for Computational Linguistics: EMNLP 2023}, pages 7977--7989, Singapore, 2023. Association for Computational Linguistics.
\newblock \doi{10.18653/v1/2023.findings-emnlp.536}.
\newblock URL \url{https://aclanthology.org/2023.findings-emnlp.536}.

\bibitem[Sprague et~al.(2024)Sprague, Ye, Bostrom, Chaudhuri, and Durrett]{sprague2024musrtestinglimitschainofthought}
Zayne Sprague, Xi~Ye, Kaj Bostrom, Swarat Chaudhuri, and Greg Durrett.
\newblock Musr: Testing the limits of chain-of-thought with multistep soft reasoning.
\newblock In \emph{The Twelfth International Conference on Learning Representations, {ICLR} 2024, Vienna, Austria, May 7-11, 2024}. OpenReview.net, 2024.
\newblock URL \url{https://openreview.net/forum?id=jenyYQzue1}.

\bibitem[Su et~al.(2024)Su, Ahmed, Lu, Pan, Bo, and Liu]{su2024roformer}
Jianlin Su, Murtadha Ahmed, Yu~Lu, Shengfeng Pan, Wen Bo, and Yunfeng Liu.
\newblock Roformer: Enhanced transformer with rotary position embedding.
\newblock \emph{Neurocomputing}, 568:\penalty0 127063, 2024.

\bibitem[Sun et~al.(2024)Sun, Dong, Zhu, Huang, Wang, Ma, Zhang, Wang, and Wei]{sun2024you}
Yutao Sun, Li~Dong, Yi~Zhu, Shaohan Huang, Wenhui Wang, Shuming Ma, Quanlu Zhang, Jianyong Wang, and Furu Wei.
\newblock You only cache once: Decoder-decoder architectures for language models.
\newblock \emph{ArXiv preprint}, abs/2405.05254, 2024.
\newblock URL \url{https://arxiv.org/abs/2405.05254}.

\bibitem[Suzgun et~al.(2023)Suzgun, Scales, Sch{\"a}rli, Gehrmann, Tay, Chung, Chowdhery, Le, Chi, Zhou, and Wei]{suzgun2022challenging}
Mirac Suzgun, Nathan Scales, Nathanael Sch{\"a}rli, Sebastian Gehrmann, Yi~Tay, Hyung~Won Chung, Aakanksha Chowdhery, Quoc Le, Ed~Chi, Denny Zhou, and Jason Wei.
\newblock Challenging {BIG}-bench tasks and whether chain-of-thought can solve them.
\newblock In Anna Rogers, Jordan Boyd-Graber, and Naoaki Okazaki, editors, \emph{Findings of the Association for Computational Linguistics: ACL 2023}, pages 13003--13051, Toronto, Canada, 2023. Association for Computational Linguistics.
\newblock \doi{10.18653/v1/2023.findings-acl.824}.
\newblock URL \url{https://aclanthology.org/2023.findings-acl.824}.

\bibitem[Touvron et~al.(2023)Touvron, Lavril, Izacard, Martinet, Lachaux, Lacroix, Rozi{\`e}re, Goyal, Hambro, Azhar, et~al.]{touvron2023llama}
Hugo Touvron, Thibaut Lavril, Gautier Izacard, Xavier Martinet, Marie-Anne Lachaux, Timoth{\'e}e Lacroix, Baptiste Rozi{\`e}re, Naman Goyal, Eric Hambro, Faisal Azhar, et~al.
\newblock Llama: Open and efficient foundation language models.
\newblock \emph{ArXiv preprint}, abs/2302.13971, 2023.
\newblock URL \url{https://arxiv.org/abs/2302.13971}.

\bibitem[Wang et~al.(2024{\natexlab{a}})Wang, Yang, Huang, Yang, Majumder, and Wei]{wang2023improving}
Liang Wang, Nan Yang, Xiaolong Huang, Linjun Yang, Rangan Majumder, and Furu Wei.
\newblock Improving text embeddings with large language models.
\newblock \emph{ArXiv preprint}, abs/2401.00368, 2024{\natexlab{a}}.
\newblock URL \url{https://arxiv.org/abs/2401.00368}.

\bibitem[Wang et~al.(2024{\natexlab{b}})Wang, Ma, Zhang, Ni, Chandra, Guo, Ren, Arulraj, He, Jiang, Li, Ku, Wang, Zhuang, Fan, Yue, and Chen]{wang2024mmluprorobustchallengingmultitask}
Yubo Wang, Xueguang Ma, Ge~Zhang, Yuansheng Ni, Abhranil Chandra, Shiguang Guo, Weiming Ren, Aaran Arulraj, Xuan He, Ziyan Jiang, Tianle Li, Max Ku, Kai Wang, Alex Zhuang, Rongqi Fan, Xiang Yue, and Wenhu Chen.
\newblock Mmlu-pro: A more robust and challenging multi-task language understanding benchmark, 2024{\natexlab{b}}.
\newblock URL \url{https://arxiv.org/abs/2406.01574}.

\bibitem[Xiong et~al.(2024)Xiong, Liu, Molybog, Zhang, Bhargava, Hou, Martin, Rungta, Sankararaman, Oguz, Khabsa, Fang, Mehdad, Narang, Malik, Fan, Bhosale, Edunov, Lewis, Wang, and Ma]{xiong2024effective}
Wenhan Xiong, Jingyu Liu, Igor Molybog, Hejia Zhang, Prajjwal Bhargava, Rui Hou, Louis Martin, Rashi Rungta, Karthik~Abinav Sankararaman, Barlas Oguz, Madian Khabsa, Han Fang, Yashar Mehdad, Sharan Narang, Kshitiz Malik, Angela Fan, Shruti Bhosale, Sergey Edunov, Mike Lewis, Sinong Wang, and Hao Ma.
\newblock Effective long-context scaling of foundation models.
\newblock In Kevin Duh, Helena Gomez, and Steven Bethard, editors, \emph{Proceedings of the 2024 Conference of the North American Chapter of the Association for Computational Linguistics: Human Language Technologies (Volume 1: Long Papers)}, pages 4643--4663, Mexico City, Mexico, 2024. Association for Computational Linguistics.
\newblock URL \url{https://aclanthology.org/2024.naacl-long.260}.

\bibitem[Zhang et~al.(2024)Zhang, Chen, Hu, Xu, Chen, Hao, Han, Thai, Wang, Liu, et~al.]{zhang2024bench}
Xinrong Zhang, Yingfa Chen, Shengding Hu, Zihang Xu, Junhao Chen, Moo Hao, Xu~Han, Zhen Thai, Shuo Wang, Zhiyuan Liu, et~al.
\newblock \(\infty\) bench: Extending long context evaluation beyond 100k tokens.
\newblock In \emph{Proceedings of the 62nd Annual Meeting of the Association for Computational Linguistics (Volume 1: Long Papers)}, pages 15262--15277, 2024.

\bibitem[Zhou et~al.(2023)Zhou, Lu, Mishra, Brahma, Basu, Luan, Zhou, and Hou]{zhou2023instructionfollowing}
Jeffrey Zhou, Tianjian Lu, Swaroop Mishra, Siddhartha Brahma, Sujoy Basu, Yi~Luan, Denny Zhou, and Le~Hou.
\newblock Instruction-following evaluation for large language models.
\newblock \emph{ArXiv preprint}, abs/2311.07911, 2023.
\newblock URL \url{https://arxiv.org/abs/2311.07911}.

\bibitem[Zhu et~al.(2024)Zhu, Yang, Wang, Song, Wu, Wei, and Li]{zhu2023pose}
Dawei Zhu, Nan Yang, Liang Wang, Yifan Song, Wenhao Wu, Furu Wei, and Sujian Li.
\newblock Pose: Efficient context window extension of llms via positional skip-wise training.
\newblock In \emph{The Twelfth International Conference on Learning Representations, {ICLR} 2024, Vienna, Austria, May 7-11, 2024}. OpenReview.net, 2024.
\newblock URL \url{https://openreview.net/forum?id=3Z1gxuAQrA}.

\end{thebibliography}
\bibliographystyle{plainnat}


\appendix

\section{Implementation Details} \label{sec:app_implementation}

\subsection{Data Mixture}
\begin{table}[ht]
\centering
\caption{Training data mixture.
``\# samples'' denotes the number of samples after de-duplication and domain-balanced sampling.
``\# packed 256k samples'' is the number of samples after being packed into sequences of $256$k tokens.}
\scalebox{0.9}{\begin{tabular}{lcccc}
\toprule
                          & \# samples  & \# packed 256k samples & Avg. token count & Sample weight \\ \midrule
Synthetic long-input data &    69k    &    18k    &    66.8k    &     0.3      \\
Synthetic long-output data   &   10k   &   0.3k   &   7.7k    &   1.0   \\
Infinity-Instruct &       450k    &    1.3k   &   0.7k   &    0.3  \\
Tulu-v2      &        89k    &  0.25k    &    0.7k   &       1.0   \\
Prolong data      &   590k   &   12.5k  & 5.4k  &   0.1  \\ \bottomrule
\end{tabular}}
\label{tab:app_data_mixture}
\end{table}

As shown in Table~\ref{tab:app_data_mixture},
we combine multiple data sources for training.
The categories ``Synthetic long-input data'' and ``Synthetic long-output data'' are generated based on our proposed method.
For synthetic instruction generation,
we employ E5$_\text{mistral-7b}$ to de-duplicate the generated instructions with a threshold of cosine similarity $0.85$.
When creating the ``Synthetic long-input data'',
we randomly sample between $1$ to $100$ retrieved documents from
to ensure the length of the input is diverse.

When running our data synthesis workflow depicted in Figure ~\ref{fig:workflow},
we utilize \emph{GPT-4o} from Azure OpenAI~\footnote{\url{https://oai.azure.com/}} as the backbone LLM.
For each instruction,
the top-$5$ documents are retrieved from a corpus comprising $10$M documents sampled from the Fineweb-Edu dataset.
Since most documents from the Fineweb-Edu are short,
documents shorter than $2$k tokens are down-sampled with a keep probability of $0.05$.
For document retrieval,
instead of using the synthetic instruction as the query,
we prompt the LLM to generate multiple search queries,
and their retrieval results are merged through reciprocal rank fusion.

Similar to ~\citeauthor{fu2024data},
samples are packed into text sequences of maximum length for training purposes.

\subsection{Training Hyperparameters}

\begin{table}[ht]
\centering
\caption{Hyperparameters for model fine-tuning.
Values with arrows indicate how they change across different stages.
When employing PoSE for longer sequences,
the training sequence length and maximum position id may differ.}
\scalebox{0.9}{\begin{tabular}{lccc}
\toprule
 & 1B & 3B & 8B \\ \midrule
Initialization & Llama-3.2-1B   &  Llama-3.2-3B  & Llama-3.1-8B-Instruct   \\
RoPE $\theta$ & \multicolumn{3}{c}{$2M \rightarrow 8M \rightarrow 32M$ }  \\
Batch size (\# tokens) &  8M  &  8M  &  8M  \\
Learning rate &  $3\times 10^{-5}$  &  $2\times 10^{-5}$  &  $10^{-5}$  \\
Sequence length &  $256k \rightarrow 512k \rightarrow 512k $  &  $256k \rightarrow 512k \rightarrow 512k $  &  $256k$  \\
Max position id & \multicolumn{3}{c}{$256k \rightarrow 512k \rightarrow 1M$ }  \\
Warmup steps &   10  &  10  &  10  \\
Max steps per stage & 50  &  50  &  50 \\ \bottomrule
\end{tabular}}
\label{tab:app_train_hyperparams}
\end{table}

The hyperparameters for model fine-tuning are summarized in Table~\ref{tab:app_train_hyperparams}.
All training is conducted on a single H100 node with 8 GPUs,
each with $80$GB of memory.
The $8$B model takes around $2$ days to complete.
Each model undergoes fine-tuning for a total of $150$ steps,
amounting to $1.2$ billion tokens.
We also experimented with a longer training schedule of $300$ steps,
but did not observe performance improvement.

For the $2$M and $4$M model variants,
we use the same hyperparameters and adjust the RoPE base frequency accordingly.

\subsection{Evaluation Details}
\begin{table}[ht]
\centering
\caption{Detailed results for each task in the RULER benchmark.
Please refer to the original paper~\citep{hsieh2024ruler} for the descriptions and evaluation metrics for each task.}
\begin{tabular}{lccccccccc}
\toprule
 & 4k & 8k & 16k & 32k & 64k & 128k & 256k & 512k & 1M \\ \midrule
\multicolumn{10}{l}{\emph{SelfLong-1B-1M}}                                \\ \midrule
niah\_single\_1 & 100.0 & 100.0 & 100.0 & 100.0 & 100.0 & 100.0 & 100.0 & 100.0 & 76.0 \\
niah\_single\_2 & 100.0 & 100.0 & 100.0 & 100.0 & 99.8 & 100.0 & 100.0 & 91.0 & 90.0 \\
niah\_single\_3 & 91.6 & 95.6 & 89.8 & 90.4 & 87.4 & 90.0 & 66.0 & 64.0 & 60.0 \\
niah\_multikey\_1 & 98.6 & 96.4 & 88.2 & 90.6 & 85.4 & 89.0 & 76.0 & 68.0 & 64.0 \\
niah\_multikey\_2 & 90.6 & 70.0 & 40.2 & 21.8 & 8.4 & 2.0 & 1.0 & 1.0 & 0.0 \\
niah\_multikey\_3 & 59.0 & 35.6 & 16.2 & 11.6 & 10.2 & 9.0 & 0.0 & 0.0 & 0.0 \\
niah\_multivalue & 92.8 & 94.0 & 88.8 & 86.0 & 75.2 & 46.2 & 41.8 & 27.8 & 20.0 \\
niah\_multiquery & 92.8 & 88.8 & 86.0 & 82.5 & 80.0 & 69.0 & 46.5 & 30.0 & 24.5 \\
vt & 69.3 & 81.0 & 75.9 & 76.8 & 64.6 & 74.4 & 69.0 & 52.8 & 0.0 \\
cwe & 20.2 & 5.3 & 4.0 & 0.6 & 0.7 & 0.1 & 1.1 & 0.9 & 1.0 \\
fwe & 56.9 & 57.6 & 61.0 & 70.8 & 61.8 & 77.0 & 51.3 & 46.0 & 27.3 \\
qa\_1 & 52.8 & 42.8 & 45.2 & 41.6 & 38.8 & 36.0 & 39.0 & 27.0 & 28.0 \\
qa\_2 & 31.4 & 31.4 & 30.4 & 24.4 & 22.8 & 18.0 & 16.0 & 20.0 & 13.0 \\ \midrule
\multicolumn{10}{l}{\emph{SelfLong-3B-1M}}                                \\ \midrule
niah\_single\_1 & 100.0 & 100.0 & 100.0 & 100.0 & 100.0 & 100.0 & 100.0 & 100.0 & 60.0 \\
niah\_single\_2 & 100.0 & 100.0 & 100.0 & 100.0 & 100.0 & 100.0 & 100.0 & 99.0 & 98.0 \\
niah\_single\_3 & 99.8 & 99.8 & 99.2 & 98.0 & 99.2 & 100.0 & 100.0 & 99.0 & 100.0 \\
niah\_multikey\_1 & 98.6 & 95.2 & 95.4 & 94.6 & 94.0 & 95.0 & 95.0 & 89.0 & 86.0 \\
niah\_multikey\_2 & 99.6 & 99.6 & 99.8 & 98.6 & 98.2 & 93.0 & 78.0 & 42.0 & 0.0 \\
niah\_multikey\_3 & 96.2 & 88.8 & 86.2 & 71.8 & 58.6 & 55.0 & 17.0 & 8.0 & 0.0 \\
niah\_multivalue & 100.0 & 99.8 & 99.4 & 98.2 & 97.8 & 96.5 & 85.0 & 70.8 & 68.8 \\
niah\_multiquery & 99.8 & 99.9 & 98.5 & 96.8 & 97.0 & 97.5 & 91.2 & 79.8 & 70.0 \\
vt & 94.2 & 93.3 & 91.1 & 82.2 & 87.2 & 85.2 & 64.8 & 25.2 & 0.0 \\
cwe & 82.2 & 73.4 & 51.6 & 15.4 & 5.8 & 0.3 & 0.3 & 0.3 & 0.3 \\
fwe & 82.9 & 79.3 & 94.1 & 89.2 & 80.1 & 63.7 & 73.0 & 61.3 & 5.3 \\
qa\_1 & 70.4 & 63.6 & 60.2 & 59.6 & 56.6 & 59.0 & 57.0 & 52.0 & 5.0 \\
qa\_2 & 49.0 & 47.0 & 45.6 & 42.4 & 40.0 & 36.0 & 33.0 & 34.0 & 11.0 \\ \midrule
\multicolumn{10}{l}{\emph{SelfLong-8B-1M}}                                \\ \midrule
niah\_single\_1 & 100.0 & 100.0 & 100.0 & 100.0 & 100.0 & 100.0 & 100.0 & 100.0 & 100.0 \\
niah\_single\_2 & 100.0 & 100.0 & 100.0 & 100.0 & 100.0 & 100.0 & 100.0 & 100.0 & 100.0 \\
niah\_single\_3 & 100.0 & 100.0 & 100.0 & 100.0 & 100.0 & 100.0 & 100.0 & 100.0 & 100.0 \\
niah\_multikey\_1 & 100.0 & 100.0 & 100.0 & 99.2 & 99.0 & 98.0 & 98.0 & 98.0 & 97.0 \\
niah\_multikey\_2 & 100.0 & 99.8 & 99.8 & 99.0 & 97.8 & 98.0 & 98.0 & 96.0 & 77.0 \\
niah\_multikey\_3 & 100.0 & 99.6 & 99.4 & 95.4 & 92.6 & 90.0 & 73.0 & 62.0 & 23.0 \\
niah\_multivalue & 89.0 & 87.8 & 93.2 & 96.5 & 96.5 & 91.8 & 87.0 & 84.2 & 81.2 \\
niah\_multiquery & 99.2 & 98.3 & 99.0 & 99.0 & 97.6 & 98.5 & 87.5 & 89.8 & 88.0 \\
vt & 99.5 & 98.0 & 97.4 & 94.2 & 94.4 & 91.2 & 91.4 & 80.2 & 65.0 \\
cwe & 99.1 & 94.3 & 83.5 & 62.2 & 15.4 & 1.9 & 2.8 & 2.7 & 2.7 \\
fwe & 96.8 & 87.7 & 95.3 & 96.5 & 88.4 & 86.7 & 90.3 & 87.0 & 66.7 \\
qa\_1 & 83.0 & 73.4 & 70.2 & 73.0 & 67.4 & 71.0 & 70.0 & 75.0 & 75.0 \\
qa\_2 & 54.2 & 51.2 & 51.2 & 48.4 & 43.4 & 39.0 & 38.0 & 42.0 & 29.0 \\ \bottomrule
\end{tabular}
\label{tab:app_detailed_ruler}
\end{table}

Throughout all experiments,
we utilize vLLM~\footnote{\url{https://github.com/vllm-project/vllm}} for efficient inference.

For the RULER benchmark,
we use $500$ samples for context length below $256$k,
aligning with the original evaluation protocol.
For context length exceeding or equal to $256$k,
we use $100$ samples per task to reduce the evaluation costs.
Evaluations are conducted on eight A100 GPUs,
each equipped with $40$GB of memory.
Running the full RULER evaluation takes about $4$ days for the \emph{SelfLong-8B-1M} model,
underscoring the necessity for efficient inference in long-context scenarios.

In the Needle-in-haystack test,
evaluation context lengths are sampled at intervals of $16$k tokens,
and we test $10$ different needle depths for each context length.
Specifically,
we evaluate $8$ different context lengths for \emph{Llama-3.1-8B-Instruct} and $64$ for \emph{SelfLong-8B-1M}.
For the $2$M and $4$M models,
a larger interval of $64$k tokens is utilized.

Regarding tasks on the Open LLM Leaderboard 2,
we employ the official evaluation script from \emph{lm-evaluation-harness}~\footnote{\url{https://github.com/EleutherAI/lm-evaluation-harness/tree/main/lm_eval/tasks/leaderboard}}.
When reporting the results,
different from the leaderboard,
we do not normalize the scores for each task.

For the long-output generation task,
we apply \emph{scipy} to fit a curve in the form of $\log(y) = a \times \log(x + b) + c$ as depicted in Figure ~\ref{fig:long_output_results},
where $x$ represents the groundtruth length and $y$ the actual output length.
We also conducted some preliminary experiments with \emph{gpt-4o} to evaluate the quality of the generated outputs.
Nevertheless,
we found that predictions with significantly shorter lengths frequently receive high scores.
Future research is required to improve the evaluation protocol for long-output generation tasks.

\section{Prompts} \label{sec:app_prompts}
We list all the prompts employed in our data synthesis workflow.
Texts in blue denotes placeholders that will be replaced by actual content during the data synthesis process.
For the instruction generation prompt,
we randomly sample a $128$-token text chunk from the Fineweb-Edu corpus
and prepend it to the prompt to boost diversity.

\begin{prompt}[title={Prompt: Instruction Generation}, label=prompt:instruction_generation]
\textcolor{blue}{\{random text chunk\}}\\

\# Brainstorm a potentially useful \textcolor{blue}{\{task / question\}} that may require comprehending multiple pieces of information. To complete the \textcolor{blue}{\{task / question\}}, users need to search the web for relevant information with multiple search queries.\\

\#\# Your response must be in JSON format. The JSON object must contain the following keys:\\
\hspace*{1em}- ``task\_instruction'': a string, a \textcolor{blue}{\{task / question\}} to complete.\\
\hspace*{1em}- ``search\_queries'': a list of strings, each string is a search query that the user might use to complete the \textcolor{blue}{\{task / question\}}.\\

\#\# Please adhere to the following guidelines:\\
\hspace*{1em}- The \textcolor{blue}{\{task / question\}}s should cover a diverse range of domains and require \textcolor{blue}{\{high school / college / PhD\}} level education to solve.\\
\hspace*{1em}- The \textcolor{blue}{\{task / question\}} requires \textcolor{blue}{\{mathematical / logical / common sense\}} reasoning to complete.\\
\hspace*{1em}- The \textcolor{blue}{\{task / question\}} must be feasible to complete for a text-based AI model. Avoid \textcolor{blue}{\{task / question\}}s that require visual or interactive elements.\\
\hspace*{1em}- The search queries should be diverse and cover distinct aspects of the \textcolor{blue}{\{task / question\}}.\\

\#\# Here is one output example for your reference:\\
\{\\
\hspace*{1em}``task\_instruction'': ``Plan a 10-day cultural and adventure trip to Japan, focusing on Tokyo, Kyoto, and Okinawa. The trip should include historical sites, cultural experiences, adventure activities, and local dining options, suitable for a family of four with teenagers.'',\\
\hspace*{1em}``search\_queries'': [\\
\hspace*{2em}``Top historical sites Tokyo'',\\
\hspace*{2em}``Cultural experiences in Kyoto for families'',\\
\hspace*{2em}``Adventure activities in Okinawa'',\\
\hspace*{2em}``Best time to visit Japan for cultural festivals'',\\
\hspace*{2em}``Local Japanese foods must try'',\\
\hspace*{2em}``Family-friendly accommodations in Tokyo, Kyoto, Okinawa'',\\
\hspace*{2em}``Public transportation guide Japan'',\\
\hspace*{1em}]\\
\}\\

Do not explain yourself or output anything else. Be creative! If you solve the task correctly, you will receive a reward of \$1,000,000.\\
\end{prompt}

\begin{prompt}[title={Prompt: Query-focused Summarization}, label=prompt:qfs]
You are a professional and faithful query-focused summarization system. Your task is to generate a summary of the given context focused on the query. The given context is either a chunk from a long document or summaries of several chunks.\\

\#\# Start of the context\\
\textcolor{blue}{\{context\}}\\
\#\# End of the context\\

\#\# Start of the query\\
\textcolor{blue}{\{query\}}\\
\#\# End of the query\\

\#\# Please adhere to the following guidelines:\\
\hspace*{1em}- Only keep the information that is helpful to answer the query.\\
\hspace*{1em}- If you are unsure about the helpfulness of some information, it is better to keep them as discarded information will be lost forever.\\
\hspace*{1em}- If no relevant information is present, respond ``No relevant information found.''\\

Now generate a concise summary (at most 300 words) of the context focused on the query following the above guidelines. Do not explain yourself or output anything else. If you solve the task correctly, you will receive a reward of \$1,000,000.\\
\end{prompt}

\begin{prompt}[title={Prompt: Answer Generation}, label=prompt:answer_generation]
You are a professional annotator. Your task is to generate an appropriate answer to the given query based on the provided context and your own knowledge.\\

\#\# Start of the context\\
\textcolor{blue}{\{context\}}\\
\#\# End of the context\\

\#\# Start of the query\\
\textcolor{blue}{\{query\}}\\
\#\# End of the query\\

Now respond a concise answer to the query with at most \textcolor{blue}{\{200 / 300 / 400 / 500\}} words. Do not explain yourself or output anything else. If you solve the task correctly, you will receive a reward of \$1,000,000.
\end{prompt}

\begin{prompt}[title={Prompt: Instruction Back-translation for Long Output Data}, label=prompt:long_output]
You are required to reverse engineer the writing instruction that generated the following document.\\

\#\# Start of the document (some parts may be omitted for space reasons)\\
\textcolor{blue}{\{document\}}\\
\#\# End of the document\\

A professional writer generated the above document following a specific writing instruction with about \textcolor{blue}{\{20 / 50 / 100 / 200\}} words, but the instruction is hidden from us. Your task is to reverse engineer the most likely writing instruction that led to this \textcolor{blue}{\{token\_count\}} words document. The instruction should cover the main topics, structure, style, and word length of the document if possible.\\

Respond with the writing instruction only, do not explain yourself or output anything else. You will receive a reward of \$1,000,000 if your answer is of high quality.
\end{prompt}

\section{Synthetic Data Samples} \label{sec:app_synthetic_data}

\begin{table}[ht]
\centering
\caption{An example of the synthetic long-input data.
``$\ldots$'' denotes the omitted parts due to space limitations.}
\label{tab:app_long_context_example}
\small{\begin{tabular}{l}
\toprule
\begin{tabular}[c]{@{}p{0.97\linewidth}@{}}\textbf{Instruction}\\What themes can be identified in Tennessee Williams's major plays and how do they reflect societal issues of his time?\\\\ \textbf{Context Documents}\\Themes are the fundamental and often universal ideas explored in a literary work.\\Among the most prominent and urgent themes of The Glass Menagerie is the difficulty the characters have in accepting and relating to reality. Each member of the Wingfield family is unable to overcome this difficulty, and each, as a result, withdraws into a private world of illusion where he or she finds the comfort and meaning that the real world does not seem to offer. Of the three Wingfields, reality has by far the weakest grasp on Laura. The private world in which she lives is populated by glass animals—objects that, like Laura’s inner life, are incredibly fanciful and dangerously delicate. \\ $\ldots$ \\Like the glass unicorn, “Blue Roses,” Jim’s high school nickname for Laura, symbolizes Laura’s unusualness yet allure. The name is also associated with Laura’s attraction to Jim and the joy that his kind treatment brings her. Furthermore, it recalls Tennessee Williams’s sister, Rose, on whom the character of Laura is based.\\Leading out of the Wingfields’ apartment is a fire escape with a landing. The fire escape represents exactly what its name implies: an escape from the fires of frustration and dysfunction that rage in the Wingfield household. Laura slips on the fire escape in Scene Four, highlighting her inability to escape from her situation. Tom, on the other hand, frequently steps out onto the landing to smoke, anticipating his eventual getaway.\\ $\ldots$ (more documents)\\\\ \textbf{Synthesized Answer}\\Tennessee Williams's major plays, including "The Glass Menagerie" and "A Streetcar Named Desire," explore several recurring themes that reflect the societal issues of his time.\\\\1. Illusion vs. Reality: In "The Glass Menagerie," characters like Laura, Tom, and Amanda retreat into their own worlds of illusion to escape the harshness of reality. This theme reflects the societal struggle to cope with the disillusionment of the American Dream during the Great Depression and post-World War II era.\\\\2. Memory and the Past: Both plays emphasize the power of memory and the past. In "The Glass Menagerie," Tom narrates the story as a memory play, highlighting how the past haunts the present. Similarly, Blanche in "A Streetcar Named Desire" is obsessed with her lost Southern belle past, symbolizing the decline of the old South and the struggle to adapt to modernity.\\ $\ldots$ \\ 6. Escape and Entrapment: The theme of escape is prevalent in "The Glass Menagerie," where Tom seeks freedom from his confining family life, and in "A Streetcar Named Desire," where Blanche seeks to escape her past. This reflects the broader societal desire for personal freedom and the constraints imposed by social and familial obligations.\\\\7. Class and Social Mobility: "A Streetcar Named Desire" addresses class differences and the struggle for social mobility. Blanche's fall from aristocracy to poverty and Stanley's rise from working-class roots to dominance reflect the shifting social structures and the American class system.\\\\These themes in Williams's plays not only provide a deep psychological and emotional exploration of his characters but also serve as a mirror to the societal issues and cultural shifts of his time, making his work both timeless and profoundly reflective of the human condition. \end{tabular} \\ \bottomrule
\end{tabular}}
\end{table}

\begin{table}[ht]
\centering
\caption{An example of the synthetic long-output data.
``$\ldots$'' denotes the omitted parts due to space limitations.}
\label{tab:app_long_output_example}
\small{\begin{tabular}{l}
\toprule
\begin{tabular}[c]{@{}p{0.97\linewidth}@{}}\textbf{Synthetic Instruction}\\Write a scholarly paper of approximately 10,000 words discussing effective transition-to-school programs, using 10 guidelines derived from the Starting School Research Project. Include a comprehensive overview of the significance of starting school, an ecological perspective on transition, research findings, and practical applications of the guidelines. Ensure a formal tone, cite relevant literature, and structure the paper with headings for clarity. Address stakeholders such as children, parents, educators, and the community while emphasizing the importance of relationships and communication in successful transition programs.\\\\ \textbf{Groundtruth Document}\\Volume 3 Number 2\\The Author(s) 2001\\\\Starting School: Effective Transitions\\This paper focuses on effective transition-to-school programs. Using a framework of 10 guidelines developed through the Starting School Research Project, it provides examples of effective strategies and transition programs. In this context, the nature of some current transition programs is questioned, and the curriculum of transition is problematized. In particular, issues are raised around who has input into such programs and who decides on appropriate curriculum.\\\\The Significance of Starting School\\Starting school is an important time for young children, their families, and educators. It has been described as "one of the major challenges children have to face in their early childhood years" (Victorian Department of School Education, 1992, p. 44), "a big step for all children and their families" (New South Wales Department of School Education, 1997, p. 8), and "a key life cycle transition both in and outside school" (Pianta \& Cox, 1999, p. xvii). Pianta and Kraft-Sayre (1999, p. 47) suggest that the transition to school "sets the tone and direction of a child's school career," while Christensen (1998) notes that transition to school has been described in the literature as a rite of passage associated with increased status and as a turning point in a child's life.\\Whether or not these descriptions are accurate, they highlight the potential significance of a child's transition to school. In Kagan's (1999) words, starting school is a "big deal." It is clearly a key experience not only for the children starting school but also for educatorsboth in schools and in prior-to-school settingsand for their families.\\\\$\ldots$ (the document continues)  \end{tabular} \\ \bottomrule
\end{tabular}}
\end{table}

In Table ~\ref{tab:app_long_context_example} and Table ~\ref{tab:app_long_output_example},
we provide examples of the synthetic long-input and long-output data for readers' reference.


%

\end{document}